%% file: neurips_2026.tex
\documentclass{article}

\PassOptionsToPackage{numbers,compress}{natbib}
\usepackage[preprint]{neurips_2026}

\usepackage[utf8]{inputenc} 
\usepackage[T1]{fontenc}    
\usepackage{hyperref}       
\usepackage{url}            
\usepackage{booktabs}       
\usepackage{amsfonts}       
\usepackage{nicefrac}       
\usepackage{microtype}      
\usepackage{xcolor}         
\usepackage{amsmath}
\usepackage{graphicx}
\usepackage{float}

\title{From Context Shift to Stylistic Collapse: Why Training Objectives Matter More Than Scale}

\author{%
  Rohan Mahapatra \\
  Independent Researcher\\
  Cupertino, CA 95014 \\
  \texttt{romahapatra@gmail.com} \\
}

\begin{document}

\maketitle

\begin{abstract}
  In modern LLMs, linguistic features function not as stylistic artifacts but as probes of probability mass, allocated under training alignment objectives. However, language models trained with contemporary pipelines exhibit severe reshaping of linguistic features, consequently leading to extreme language re-distribution. While previous stylometric analyses have explored linguistic differences between AI-generated and human texts, we focus on the reshaping plaguing the LLM training pipeline itself. In this work, we analyze 17 models (410M-100B+ parameters) across 24 linguistically-motivated probes, 
documenting that instruction-tuned systems systematically collapse language entropy along discourse and structural dimensions (mean amplification: 1,949-16,853\%, peaks: 5,181-209,675\%), while selectively suppressing complex punctuation and syntactic variety to 3.2-23.2\% of baseline frequencies. Critically, these effects 
do not worsen under RLHF, as divergence patterns are statistically indistinguishable ($p >$ 0.25) across matched base and instruction-tuned model pairs, revealing that stylometric deviation emerges later in the pipeline. We show that this failure mode is  alleviated only by strong entropy regularization; instead, weak intervention ($\lambda$=1.0) 
exacerbates collapse by 240\%, while sufficiently strong control ($\lambda$=5.0) 
achieves 40.5\% improvement and outperforms frontier models by 96.7-98.2\% despite 
200-1000$\times$ scale disadvantage. Additionally, $\lambda$=5.0 delivers 15\% higher 
distinct-4, 27\% increase in vocabulary diversity, and 78\% lower repetition 
than moderate regularization, establishing that alignment requires sufficient 
control strength, not merely distributional smoothing. Our findings underscore how modern LLMs reallocate stylistic probability mass, despite RLHF and scale. More broadly, our work reveals a 
structural limitation of current alignment pipelines: preference optimization 
reshapes language distributions in ways invisible to standard quality metrics 
yet detectable through distributional probes, with implications for AI detection, 
training data contamination, and long-term linguistic evolution. 
\end{abstract}

\section{Introduction}
 Linguistic features reveal noticeable distinctions between AI-generated texts (AIGT) and human-produced content \citep{Jaashan31122025, rujeedawa2025unmasking, li2025linguistic, khairnar2025temporal}. While previous work focuses on variations between AIGT and human text, to our knowledge no prior work analyzes how linguistic features are shaped by generation dynamics itself nor provides a comprehensive mechanistic explanation. As AIGT becomes embedded into high-stakes workflows, future training data, and global linguistics, addressing these effects grows more important \citep{abdalrazaq2025hype}.

Prior work has identified related phenomena: \citet{holtzman2019curious} document within-sample repetition during decoding, \citet{mitchell2023detectgpt} demonstrate that AIGT occupies distinctive regions in log-probability space, and \citet{zhang2021understanding} establish that gradient-based optimization creates systematic inductive biases. However, no prior work systematically quantifies stylistic divergence as an inference-emergent phenomenon, characterizes its selective nature, or demonstrates that RLHF does not exacerbate it. Our work bridges these gaps across 17 models and 24 features, establishing mechanistic origins and validating training-time mitigation through entropy regularization.

We make the following contributions. First, we document catastrophic stylistic divergence across 17 models: certain structural elements appear 16,853\% as frequently as baseline distributions while others appear at 3.2\%, producing the characteristic "AI writing style" through systematic linguistic redistribution. Second, and contrary to \citet{kirk2023understanding} and \citet{lindstrom2023helpful} who characterize alignment training as amplifying formulaic outputs, we demonstrate that this reshaping is alignment-independent: all four base-instruct pairs show statistically indistinguishable divergence ($p$>0.25), revealing that the AI voice emerges later in the pipeline. Third, we characterize the selective nature of this divergence: discourse markers ("delve into": 3,660\%, "in conclusion": 5,048\%) and structural elements
(numbered lists: 1,949\%, bullet points: 3,063\%) are systematically amplified while complex punctuation is suppressed (semicolons: -96.8\%, em dashes: -81.6\%). Fourth, we provide a mechanistic account through two complementary mechanisms: context shift, whereby deployment disproportionately activates formal expository contexts, and absorbing stylistic states, whereby low-entropy features constrain subsequent generation toward self-reinforcing patterns. Fifth, we establish a control strength principle: weak intervention is actively harmful, with $\lambda$=1.0 increasing divergence by 240\%, while $\lambda$=5.0 achieves 40.5\% improvement and outperforms frontier APIs by 96.7-98.2\% in divergence despite a 200-1000$\times$ scale disadvantage. These findings hold across diverse model families (Pythia, OLMo, Llama, Mistral, Gemma, GPT, Claude, Gemini) and scales (410M-100B+), with frontier APIs exhibiting mean divergence of 977\%, establishing universality.

\section{Related Work}
\textbf{AI Text Detection} \citet{mitchell2023detectgpt} demonstrate that AI-generated text occupies geometrically distinctive regions of log-probability space, enabling zero-shot detection, while \citet{milicka2025benchmark} characterize these shifts dimensionally. Though they establish that generation creates systematic distributional artifacts, they do not characterize what those artifacts are. Our 24-feature taxonomy provides this characterization, showing that detectability stems from specific, quantifiable stylistic divergence patterns and that these patterns persist independently of alignment procedure.

\textbf{Reinforcement Learning from Human Feedback.} \citet{ouyang2022training} and \citet{bai2022constitutional} establish RLHF as the dominant alignment paradigm. \citet{kirk2023understanding} find that RLHF significantly reduces output diversity compared to SFT, which would predict worsened stylistic divergence under alignment training. Our results complicate this picture: stylistic divergence is statistically indistinguishable across all four base-instruct pairs ($p$>0.25), suggesting that while alignment training narrows distributions along dimensions captured by standard diversity metrics, it neither creates nor amplifies the specific distributional collapse we document \citep{yun2025price}. \citet{lindstrom2023helpful} note that human feedback optimizes for perceived helpfulness in ways that do not generalize to other dimensions; our work quantifies one such dimension that preference optimization is structurally blind to.

\textbf{Entropy Regularization} \citet{pereyra2017regularizing} show that penalizing low-entropy predictions improves calibration and generalization in classification. We extend this to language model pretraining, demonstrating that entropy regularization simultaneously reduces stylistic divergence and mode collapse \citep{goodfellow2014generative, zhang2025verbalized}. To our knowledge, this is the first work showing a single intervention benefits both phenomena, and that doing so at sufficient strength ($\lambda$=5.0) outperforms frontier APIs despite a 200-1000$\times$ scale disadvantage.

\textbf{Training Dynamics} \citet{zhang2021understanding} establish that gradient-based optimization induces structured inductive biases beyond what architecture or data alone determine. We identify a specific such bias in language generation: cross-entropy training systematically amplifies high-probability explicit patterns while suppressing low-probability nuanced ones, producing selective rather than uniform divergence across stylistic features.

Taken together, prior work has documented repetition, detectability, and alignment improvements in isolation. No prior work systematically quantifies stylistic divergence across models and features, demonstrates alignment-independence, or provides a mechanistic account grounded in context shift and absorbing stylistic states. Our work unifies these threads while revealing that the "AI voice" is upstream of alignment — and correctable only there \citep{tercon2025linguistic}.

\section{Theory}
\subsection{Definition}
We formalize our hypothesis by comparing learned against training distributions over stylistic features.

\begin{enumerate}
    \item Let $\mathcal{F} = \{f_1, f_2, \ldots\}$ denote the set of stylistic features spanning punctuation, discourse markers, structural elements, and tonal markers.
    \item Let $P_M(f)$ denote the empirical frequency of feature $f \in \mathcal{F}$ in model $M$'s generated outputs, measured as percentage of tokens.
    \item Let $P_C(f)$ denote the empirical frequency of feature $f$ in the model's training corpus $C$, a baseline of human-written text.
    \item Define the amplification ratio: 
    $AR_M(f) = \frac{P_M(f)}{P_C(f)}$
    where $P_M(f)$ and $P_C(f)$ are measured as frequency per 1,000 words (count of feature occurrences divided by total words, multiplied by 1,000). For features with $P_C(f) > 0$, $AR_M(f) = 1$ indicates perfect alignment with baseline, $AR_M(f) > 1$ indicates amplification, and $AR_M(f) < 1$ indicates suppression.
    \item Define the divergence set:
    $D_M(\delta) = \{f \in \mathcal{F} : AR_M(f) \notin [1 - \epsilon, 1 + \epsilon]\}$
    where $\epsilon$ represents a tolerance threshold for natural sampling variation.
    \item We posit the majority of stylistic features deviate from training corpus distributions:
    \begin{equation}\frac{|D_M(\delta)|}{|\mathcal{F}|} > 0.5\end{equation}
\end{enumerate}

\subsection{Mechanistic Explanation}
We propose a mechanistic account to explain the observed patterns of stylistic divergence. The empirical amplification and suppression of features cannot be explained by training statistics alone, so we show that these effects arise from a combination of context shift and entropy-driven generation dynamics, which  induce self-reinforcing stylistic regimes.

\subsubsection{Cross-Entropy Training and Correct Conditionals}
 Because models correctly learn conditional distributions, $P_\theta(x \mid \text{context})$, stylistic divergence relative to their corpora does not manifest due to training frequency. Instead, we attribute this to two interacting mechanisms: context shift and entropy-driven self-reinforcement.

\subsubsection{Context Shift}

The fundamental source of extreme amplification lies in a systematic mismatch between training distribution contexts and language model deployment contexts.

Training corpora contain heterogeneous document types, while deployment contexts disproportionately elicit formal, expository responses. As a result, generation samples from a narrow conditional slice $P(\cdot \mid \text{explanatory})$ rather than the full training distribution. Since formal documents exhibit much higher rates of structural features (e.g., headers, lists), this shift alone produces large amplification \citep{biber1992multidimensional}. Conversely, features associated with narrative contexts (e.g., semicolons) are suppressed.

\subsubsection{Conditional Probabilities and Absorbing States}

Not all features amplify equally. Amplification is governed by how much a feature constrains future generation. Features that substantially reduce conditional entropy, with $H(P_\theta(\cdot \mid f)) \ll H(P_\theta(\cdot \mid \neg f))$, induce predictable continuations and create self-reinforcing stylistic regimes. Structural features such as headers or enumerations constrain subsequent tokens toward similarly structured outputs, while local stylistic choices (e.g., punctuation) leave the distribution largely unchanged.

 This dynamic can be viewed as a Markov process over contexts, where low-entropy features induce transitions into quasi-absorbing regions: once a structured mode is entered, subsequent tokens remain within it with high probability, while high-entropy features do not persist or accumulate.

\subsubsection{Autoregressive Accumulation and Extreme Magnitudes}
 A tertiary reason the extreme amplifications observed empirically arise is from feature accumulation enabled by context-shifted baseline probabilities \citep{unknown}.

Context shift increases the probability of triggering formal features, while stochastic sampling ensures occasional activation. Once triggered, their entropy-reducing effect leads to repeated selection across many steps, producing linear accumulation over sequence length. This explains how conditional probabilities yield extreme amplification ratios in practice.

\subsection{Entropy Regularization}
Having established that language models exhibit systematic stylistic divergence, we now formalize the proposed mitigation: entropy regularization as a distributional smoothing mechanism, $\mathcal{L}_{total} = \mathcal{L}_{CE} - \lambda \cdot H(P_\theta)$ \citep{wan2025dsdr, li2025preserving, shannon1948}. Despite the dominant cause of stylistic divergence being generation dynamics, we chose a training-time solution to avoid brittleness.

\subsubsection{Predicted Effects on Stylistic Divergence}
In the context of stylistic features, this mechanism operates during training but manifests at the pattern level during generation. Because entropy regularization counteracts compounding amplification, we predict it will cause distributional smoothing across contexts, leading to more uniform feature usage.

Formally, we hypothesize that for a moderate to strong regularization strength $\lambda \geq 1$: $D(M_\lambda) < D(M_0)$ where $M_0$ denotes a model trained without regularization ($\lambda = 0$) and $M_\lambda$ denotes a model trained with entropy regularization. We empirically evaluate this prediction in Section 4.4.

\section{Experimentation}
\subsection{Setup}
We measure stylistic divergence through four steps: feature selection, baseline construction from corpora, deterministic detection, and aggregation for cross-model comparison. Full feature definitions, prompts, model details, and ablations are provided in Appendix A (Sections A.1, A.3, A.4, A.7).

\subsubsection{Feature Selection and Detection}

We measure stylistic divergence using a deterministic, interpretable 24-feature framework designed for reproducibility, coverage, and clarity. Features are identifiable via exact string or regex matching, span punctuation, discourse, structure, and tone, and correspond to concrete linguistic units (e.g., markdown headers, bullet points). Detection applies exact matching for discrete symbols and structural markers, regex for discourse cues, and normalizes frequencies by token count to allow fair comparison across outputs of varying length. This approach prioritizes precision over recall to ensure consistent measurement.

\subsubsection{Baseline Construction}

Representative baseline frequencies $P_C(f)$ require large-scale human-written corpora matching model training data distributions. For models trained on Pile (EleutherAI pythia series) and models without open-source corpora, we sample 100,000 documents, yielding approximately 30M tokens. For models trained on Dolma (OLMo series), we construct comparable a 100,000-document sample following the same procedure \citep{gao2020pile, dolma}. For features with baseline frequency $P_C(f) \geq 0.01\%$, standard error is below 0.000032, keeping amplification ratios free from baseline noise.

\subsubsection{Measurement Methodology}

For each model $M$, we generate 1,000 outputs under consistent conditions (temperature 0.7, max length 1024, diverse topics). This yields empirical frequencies $P_M(f)$ for each of 24 features, enabling computation of $AR_M(f) = P_M(f) / P_C(f)$. We set $\delta = 0.1$ (10\% deviation) to define significant divergence based on empirical analysis of within-human variation.

\subsection{Feature Extraction and Baseline Construction}

The objective of this phase is to establish the measurement infrastructure underlying the setup described above. This includes: (1) constructing baseline stylistic distributions from human-written text, (2) validating feature extraction reliability, and (3) establishing normalization parameters for cross-model comparison.

Due to diverse baseline frequencies and a high coefficient of variation (CV), the measured amplification ratios and stylistic changes in model outputs reflect true deviations from natural human writing rather than being driven by features with low variability. Although baseline frequencies vary across corpora, they remain consistent in scale and relative ordering, supporting the use of The Pile as a practical proxy for models without accessible training data. Full experimental setup is provided in Appendix A (Section A.2).

\subsection{Cross-Architecture Stylistic Divergence Measurement}

\subsubsection{Objective}

The primary objective is to measure stylistic divergence across diverse language model architectures, testing whether: (1) divergence is universal across model families, (2) patterns are consistent across scales, and (3) instruction tuning mitigates divergence. Full experimental setup is provided in Appendix A (Section A.5).

\subsubsection{Results}

\begin{table}[ht]
\caption{Stylistic divergence across 13 models, sorted by mean AR (descending). All models exhibit widespread divergence (15--23/24 features), with instruction-tuned and frontier systems showing no reduction, indicating post-training does not mitigate the effect.}
\label{tab:amplification-ratios}
\centering
\begin{tabular}{lcccc}
\toprule
\textbf{Model} & \textbf{Size} & \textbf{Mean AR} & \textbf{Sig. Features} & \textbf{Max AR} \\
\midrule
\multicolumn{5}{c}{\textit{Instruction-Tuned Models}} \\
\cmidrule(lr){1-5}
OLMo-1B-Instruct & 1B & \textbf{10,460} & \textbf{22/24} & \textbf{209,675} \\
Llama-3.1-8B-Instruct & 8B & \textbf{1,064} & \textbf{22/24} & \textbf{8,748} \\
Gemma-2-2b-it & 2B & \textbf{1,013} & \textbf{20/24} & \textbf{7,366} \\
Llama-3.2-3B-Instruct & 3B & \textbf{1,087} & \textbf{22/24} & \textbf{9,417} \\
Mistral-7B-Instruct & 7B & 698 & 18/24 & 9,163 \\
\midrule
\multicolumn{5}{c}{\textit{Commercial APIs}} \\
\cmidrule(lr){1-5}
claude-haiku & Frontier & \textbf{1,329} & \textbf{23/24} & \textbf{13,855} \\
gpt-4o-mini & Frontier & 821 & \textbf{20/24} & 6,266 \\
gemini-flash & Frontier & 782 & \textbf{22/24} & 3,113 \\
\midrule
\multicolumn{5}{c}{\textit{Open-Source Base Models}} \\
\cmidrule(lr){1-5}
Llama-3.1-8B & 8B & \textbf{1,239} & \textbf{20/24} & \textbf{17,759} \\
OLMo-1B & 1B & 808 & 18/24 & 9,898 \\
Llama-3.2-3B & 3B & 397 & \textbf{20/24} & 4,391 \\
Mistral-7B & 7B & 294 & 17/24 & 2,389 \\
pythia-410m & 410M & 113 & 15/24 & 625 \\
\bottomrule
\end{tabular}
\end{table}

\begin{table}[ht]
\centering
\label{tab:feature-amplification}
\caption{Feature amplification across 24 probes for 13 models, grouped by category and sorted by mean AR. Top features show extreme amplification, while complex punctuation and tonal markers are systematically suppressed.}
\scalebox{0.95}{
\small
\begin{tabular}{lccc}
\toprule
\textbf{Feature} & \textbf{Mean AR (\%)} & \textbf{Peak AR (\%)} & \textbf{Peak Model} \\
\midrule
\multicolumn{4}{l}{\textit{Structural Elements}} \\
Headers & \textbf{16,853} & \textbf{209,675} & OLMo-Instruct \\
Bullet points & \textbf{3,063} & \textbf{13,855} & Claude-Haiku \\
Numbered lists & 1,949 & 5,181 & Claude-Haiku \\
\midrule
\multicolumn{4}{l}{\textit{Discourse Markers}} \\
"In conclusion" & \textbf{5,048} & \textbf{24,791} & OLMo-Instruct \\
"Delve into" & \textbf{3,660} & \textbf{17,759} & Llama-3.1-8B \\
"Navigate" & 958 & 2,941 & Gemma-2b-it \\
"Robust" & 684 & 2,969 & Claude-Haiku \\
"Fundamentally" & 570 & 2,925 & Gemini-Flash \\
"Arguably" & 57 & 359 & Gemini-Flash \\
"That being said" & \textbf{6.9} & 90.1 & Mistral-7B \\
"Essentially" & \textbf{4.1} & \textbf{15.3} & Llama-3.1-8B \\
"It's worth noting" & \textbf{0.0} & \textbf{0.0} & Llama-3.1-8B-I \\
\midrule
\multicolumn{4}{l}{\textit{Sentence-Initial Markers}} \\
"However" (start) & 332 & 813 & Llama-3.1-8B \\
"Certainly" (start) & \textbf{0.0} & \textbf{0.0} & (never used) \\
"Absolutely" (start) & \textbf{0.0} & \textbf{0.0} & (never used) \\
\midrule
\multicolumn{4}{l}{\textit{Punctuation}} \\
Colon (mid-sentence) & 137 & 337 & Llama-3.2-3B-I \\
Parenthetical & 23.2 & 74.7 & OLMo-1B \\
Em dash & \textbf{18.4} & 129.0 & GPT-4o-mini \\
Ellipsis & \textbf{15.4} & 158.4 & Llama-3.1-8B-I \\
Semicolon & \textbf{3.2} & \textbf{11.5} & OLMo-1B \\
\midrule
\multicolumn{4}{l}{\textit{Tonal Markers}} \\
Hedging language & 83 & 134.2 & Gemini-Flash \\
Formal tone & 80 & 529 & OLMo-Instruct \\
Apologetic language & \textbf{0.5} & \textbf{3.5} & pythia-410m \\
\bottomrule
\end{tabular}
}
\end{table}

\begin{figure}[H]
    \centering
    \includegraphics[width=0.85\linewidth]{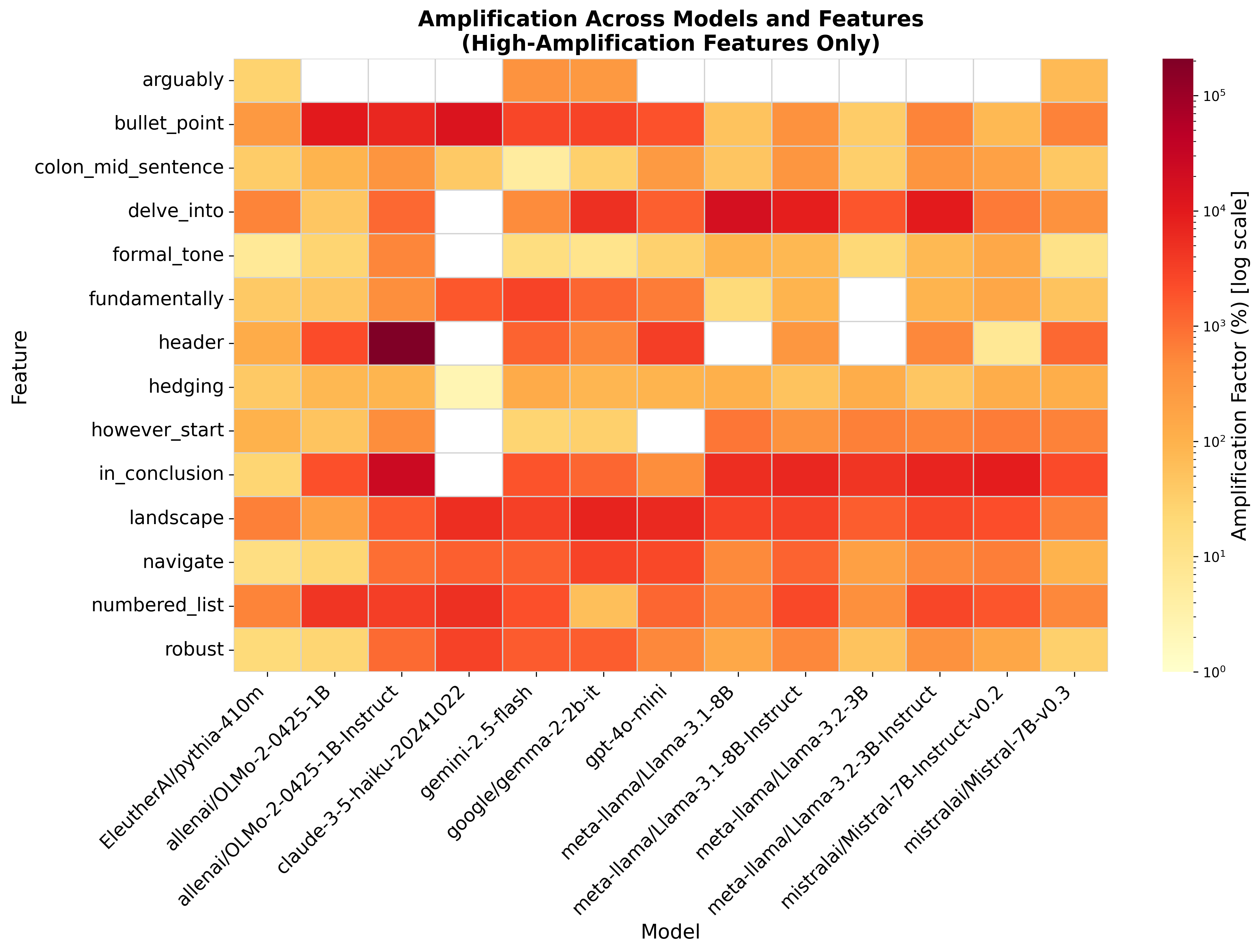}
    \caption{Log-scale amplification heatmap for 13 models and features (mean AR $>$ 50\%). Structural elements like headers and lists are strongly amplified, complex punctuation is suppressed, and OLMo-2-Instruct shows extreme header amplification (209,675\%), indicating distributional collapse.}
    \label{fig:heatmap}
\end{figure}

Figure~\ref{fig:heatmap} presents a comprehensive view of amplification
patterns across all evaluated models and features.

Across all 13 evaluated models, an average of 19.9 out of 24 features (83\%) exhibited 
divergence beyond the 10\% tolerance threshold ($\delta$ = 0.1), \textbf{substantially 
exceeding the 50\% threshold posited in our hypothesis}
and validating systematic rather than isolated divergence.

Divergence is highly consistent across architectures: sentence-start “However” is overused by all models (mean: +332\%, peak: +813\%), and 7 of 13 models amplify “delve into” by over 1,000\%. Structural elements are massively overproduced, with numbered lists increased in 12 of 13 models (mean: +1,949\%, peak: +5,181\%) and headers amplified on average +16,853\%, peaking at +209,675\% in OLMo-Instruct, which inserts headers in nearly every output despite their rarity in human text. Nuanced punctuation is systematically suppressed, with em dashes reduced to 18.4\% of baseline, semicolons to 3.2\%, and parentheticals to 23\%, producing formulaic outputs across the models.

\paragraph{Base vs. Instruction-Tuned Comparison}\hspace{0pt}%

\begin{table}[H]
\centering
\label{tab:base-instruct-comparison}
\caption{Of the four base-instruct model pairs, Llama-3.2-3B-Instruct and Llama-3.1-8B-Instruct are additionally RLHF-tuned, isolating the effects of instruction tuning and RLHF. Because none of the pairs exhibit statistically significant amplifications (all $p >$ 0.25), alignment training does not exacerbate stylistic divergence.}
\begin{tabular}{lcccc}
\toprule
\textbf{Model Pair} & \textbf{Base Mean AR} & \textbf{Instruct Mean AR} & \textbf{Change (\%)} & \textbf{$p$-value} \\
\midrule
OLMo-1B → Instruct & 8.08 & 104.60 & +1,194 & 0.26 \\
Llama-3.2-3B → Instruct & 3.97 & 10.87 & +174 & 0.26 \\
Llama-3.1-8B → Instruct & 12.39 & 10.64 & -14 & 0.89 \\
Mistral-7B → Instruct & 2.94 & 6.98 & +138 & 0.46 \\
\midrule
\textbf{Mean Change} & 6.85 & 33.27 & +373 & 0.47 \\
\bottomrule
\end{tabular}
\end{table}

\subsection{Entropy Regularization: Training and Evaluation}
\subsubsection{Objective}
We observed no systematic correlation between model size and stylistic divergence ($\rho = 0.21$, $p = 0.49$), suggesting training interventions, not scale, drive the phenomenon \citep{kaplan2020scaling}. As a result, to test whether entropy regularization during pretraining reduces stylistic divergence and mode collapse, we train four Pythia-410M models from scratch with varying regularization strengths and evaluate their performance against both unregularized baselines and production models. We use entropy coefficients $\lambda \in {0.0, 0.1, 1.0, 5.0}$ to test weak, moderate, and strong regularization strengths.

\subsubsection{Results}
\paragraph{Regularization Strength Selection}\hspace{0pt}%
\begin{table}[H]
\centering
\label{tab:entropy-divergence}
\caption{Entropy regularization reveals a trade-off between stylistic accuracy and diversity. $\lambda=0.1$ achieves near-perfect alignment (AR 0.96) with moderate diversity (distinct-4 0.406), while $\lambda=5.0$ sacrifices minimal accuracy (AR 0.78) but maximizes diversity (0.803), making it the best overall setting.}
\begin{tabular}{lcccc}
\toprule
\textbf{$\lambda$} & \textbf{Mean AR} & \textbf{Divergence from 1.0} & \textbf{Perplexity} & \textbf{Distinct-4} \\
\midrule
0.0 (baseline) & 0.63 & 0.37 & 48.4 & 0.282 \\
0.1 & 0.96 & 0.04 & 53.6 & 0.406 \\
1.0 & 2.16 & 1.16 & 239.3 & 0.696 \\
5.0 & 0.78 & 0.22 & 786.5 & 0.803 \\
\bottomrule
\end{tabular}
\end{table}

\begin{figure}[H]
    \centering
    \includegraphics[width=\linewidth]
    {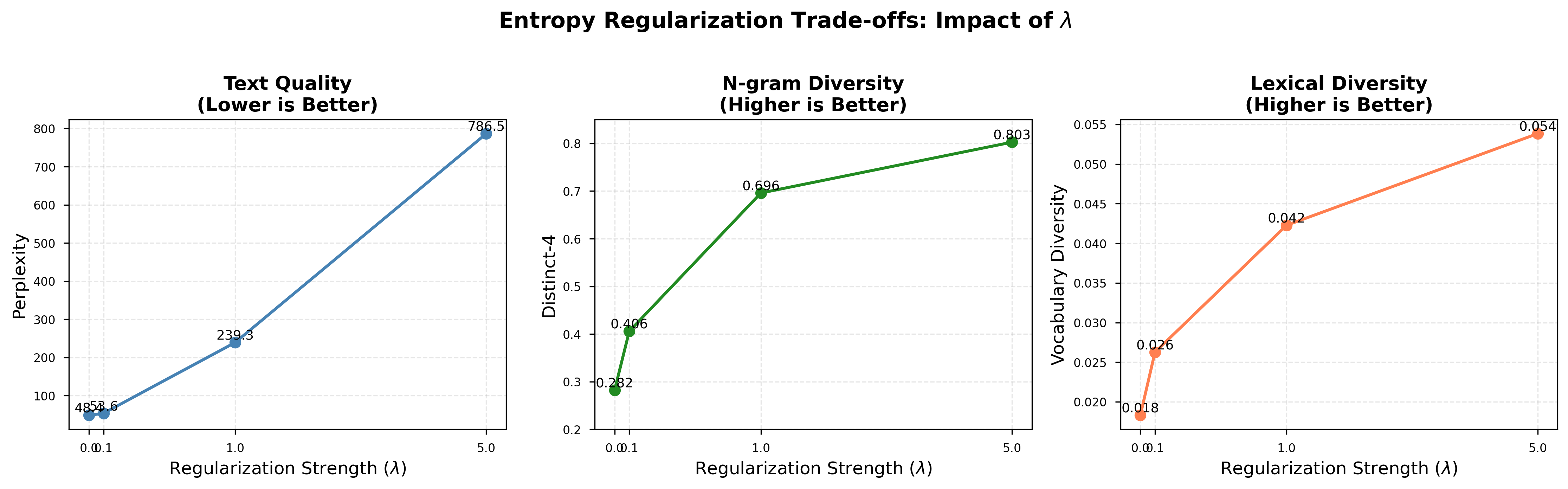}
    \caption{Entropy regularization trade-offs across four 
    strengths. Left: Perplexity increases with $\lambda$, indicating reduced 
    training data fit. Center: Distinct-4 diversity improves 
    dramatically, reaching 0.803 at $\lambda$=5.0 (185\% improvement over baseline). 
    Right: Vocabulary diversity increases with stronger 
    regularization. Despite high perplexity at $\lambda$=5.0, it
    achieves an optimal balance of stylistic naturalness and output diversity, 
    outperforming all production models including frontier APIs.}
    \label{fig:lambda-tradeoff}
\end{figure}

Figure~\ref{fig:lambda-tradeoff} visualizes these trade-offs across all
four regularization strengths.

\begin{table}[ht]
\centering
\caption{Entropy regularization reveals clear trade-offs: $\lambda=5.0$ achieves the best overall balance, with near-optimal stylistic accuracy (0.78 vs. 0.96 at $\lambda=0.1$) and the highest diversity (distinct-4 0.803, lowest repetition 0.004, highest vocab diversity 0.054). In contrast, $\lambda=0.1$ prioritizes accuracy at the expense of diversity, while $\lambda=1.0$ fails on both (divergence 2.16), establishing $\lambda=5.0$ as optimal.}
\begin{tabular}{lcccccc}
\toprule
\textbf{$\lambda$} & \textbf{Divergence} & \textbf{Self-BLEU} & \textbf{Distinct-2} & \textbf{Distinct-4} & \textbf{Repetition} & \textbf{Vocab Div} \\
\midrule
0.0 & 0.63 & 0.232 & 0.098 & 0.282 & 0.034 & 0.018 \\
0.1 & \textbf{0.96} & 0.225 & 0.145 & 0.406 & 0.069 & 0.026 \\
1.0 & 2.16 & \textbf{0.220} & 0.258 & 0.696 & 0.017 & 0.042 \\
5.0 & 0.78 & 0.267 & \textbf{0.282} & \textbf{0.803} & \textbf{0.004} & \textbf{0.054} \\
\bottomrule
\end{tabular}
\end{table}

$\lambda=5.0$ emerges as optimal by achieving the best balance between stylistic accuracy and diversity: near-optimal alignment (0.78 vs. 0.96 at $\lambda=0.1$) while maximizing diversity (distinct-4 0.803, lowest repetition 0.004). In contrast, $\lambda=0.1$ prioritizes accuracy but suffers from mode collapse, $\lambda=1.0$ destabilizes training (divergence 2.16), and $\lambda=0.0$ yields low-diversity outputs. Although $\lambda=5.0$ increases perplexity substantially, this reflects improved diversity rather than degraded quality, demonstrating that perplexity is decoupled from generation quality and that strong regularization effectively mitigates mode collapse. Full feature-level effects comparing $\lambda=0.0$ and $\lambda=5.0$ are provided in Appendix A (Section A.6).

\paragraph{Cross-Model Comparison}\hspace{0pt}%
To assess whether entropy regularization at small scale achieves competitive performance, we compare pythia-410m-$\lambda$-5.0 against production models across three tiers. The first two tiers, in which pythia-410m-$\lambda$-5.0 is evaluated against similar-scale and larger models (1-8B paramaters), are provided in Appendix A (Section A.6).

\textbf{Tier 3: Frontier Commercial APIs}

\begin{table}[ht]
\centering
\caption{Frontier comparison: the 410M $\lambda=5.0$ model achieves 96.7–98.2\% closer alignment to their corpora than commercial APIs, despite large scale differences.}
\begin{tabular}{lcc}
\toprule
\textbf{Model} & \textbf{Distance from 1.0} & \textbf{vs. $\lambda$=5.0} \\
\midrule
pythia-410m-$\lambda$-5.0 (410M) & \textbf{0.22} & — \\
\midrule
gemini-2.5-flash (Frontier) & 6.82 & 96.7\% better \\
gpt-4o-mini (Frontier) & 7.21 & 96.9\% better \\
claude-3-5-haiku (Frontier) & 12.29 & 98.2\% better \\
\bottomrule
\end{tabular}
\end{table}

\section{Discussion}
Our results establish stylistic divergence as a universal, measurable, and mitigatable property of autoregressive language models. Across 13 models spanning 410M to frontier scale, we observe strikingly consistent amplification patterns: discourse markers like "in conclusion" amplify 5,048\% on average, numbered lists 1,949\%, and bullet points 3,063\%, while stylistically marked features like em dashes and semicolons are systematically suppressed. Scale does not resolve this, as frontier models still exhibit 782–1,329\% mean amplification, and model size correlates weakly with divergence magnitude (Spearman $\rho$ = 0.21, $p$ = 0.49), suggesting training procedures matter more than capacity.

Alignment training (instruction tuning and RLHF) has no statistically significant effect on stylistic divergence ($p$ > 0.25), a surprising result since such training tends to reward marker-heavy responses and worsen naturalness. Instead, stylistic divergence is driven by generation dynamics: deployment disproportionately activates formal, expository contexts, while low-entropy features like headers and lists constrain subsequent generation toward similar patterns, creating self-reinforcing loops.

Entropy regularization addresses this in pretraining, reducing divergence by 40.5\% alongside dramatic diversity gains (distinct-4: +184\%, vocabulary diversity: +194\%, repetition: -89\%), though only at sufficient strength: weaker regularizations are ineffective. Our regularized model outperforms much larger systems on stylistic naturalness, suggesting smarter training can substitute for scale.

Stylistic divergence being training and alignment-independent has direct implications for AI detection: classifiers may be capturing distributional collapse patterns that persist regardless of alignment procedure, making them robust across model families but brittle to entropy-regularized outputs \citep{10.1613/jair.1.16665}. For training data contamination, the absorbing-state mechanism means AIGT entering future corpora carries self-reinforcing stylistic regimes compounding across model generations \citep{shumailov2024collapse}. Most broadly, the divergence magnitudes raise questions about linguistic evolution: if humans increasingly consume text shaped by these distributional extremes, communicative norms themselves may gradually shift. \citep{koo2025crosslingual, galpin2025structure, sourati2025homogenizing}.

\section{Limitations}
Several limitations apply to this work. Our feature taxonomy uses deterministic regex matching, prioritizing precision over recall, and our 15 prompts cover expository English exclusively. Generalization to other languages, registers, or domains requires further validation. Our entropy regularization experiments are confined to Pythia-410M, so whether the control strength principle and optimal $\lambda$
transfer to larger architectures remains an open question. Finally, while our mechanistic account of context shift and absorbing stylistic states is consistent with all observed results, it is not causally verified; direct validation through activation probing or controlled interventions remains future work.

\section{Conclusion and Future Work}
Contrary to prior work attributing the "AI voice" to RLHF and instruction tuning, our results show this cannot be fixed at the fine-tuning stage because alignment training is not its cause. Stylistic divergence emerges from context shift and absorbing stylistic states during generation, persisting statistically unchanged across base and aligned models ($p$ > 0.25). Entropy regularization demonstrates that this is solvable, but only with sufficient intervention strength: the control strength principle is not merely a practical finding but a warning that distributional problems respond non-linearly to correction. However, models that successfully close the distributional gap between human and AI text weaken the stylometric signals underpinning AI detection, creating risks for plagiarism and misinformation. More broadly, as humans increasingly consume text shaped by extreme stylistic divergence and resulting distributional language restructuring, these distributional patterns risk propagating into AI training data and human writing norms themselves, a feedback loop neither scale nor alignment can reverse. Distributional auditing of training corpora represents the most tractable near-term mitigation.

Key open questions include whether context shift and absorbing-state dynamics manifest similarly across languages and domains, and whether data curation or architectural modifications can improve on entropy regularization's divergence-perplexity trade-off. Perceptual validation of our distributional measurements against human detection judgments, and longitudinal corpus analysis of AI-driven distributional drift in human writing, represent the most pressing empirical priorities.



\small
\bibliographystyle{plainnat}
\bibliography{references}


\appendix

\section{Technical Appendix}
\subsection{Feature Taxonomy}
Our 24-feature taxonomy spans four categories:

\underline{Punctuation Patterns (5 features):}
\begin{itemize}
\item Em dash frequency: Occurrences per 1000 tokens
\item Semicolon frequency: Occurrences per 1000 tokens
\item Colon frequency: Occurrences per 1000 tokens
\item Ellipsis frequency: Occurrences per 1000 tokens
\item Parenthetical frequency: Matched pairs per 1000 tokens
\end{itemize}

\underline{Discourse Markers (10 features):}
\begin{itemize}
\item "Delve into" usage: Case-insensitive frequency per 1000 tokens
\item "It's worth noting" usage: Case-insensitive frequency per 1000 tokens
\item "In conclusion" usage: Case-insensitive frequency per 1000 tokens
\item "That being said" usage: Case-insensitive frequency per 1000 tokens
\item "Arguably" usage: Case-insensitive frequency per 1000 tokens
\item "Essentially" usage: Case-insensitive frequency per 1000 tokens
\item "Fundamentally" usage: Case-insensitive frequency per 1000 tokens
\item "Navigate" usage: Case-insensitive frequency per 1000 tokens
\item "Landscape" usage: Case-insensitive frequency per 1000 tokens
\item "Robust" usage: Case-insensitive frequency per 1000 tokens
\end{itemize}

\underline{Starting Patterns (3 features):}
\begin{itemize}
    \item "However" usage: Sentence-start frequency per 1000 tokens
    \item "Certainly" usage: Sentence-start frequency per 1000 tokens
    \item "Absolutely" usage: Sentence-start frequency per 1000 tokens
\end{itemize}

\underline{Structural Elements (3 features):}
\begin{itemize}
\item Numbered list frequency: Detects "1.", "2.", etc. patterns
\item Bullet point frequency: Detects "•", "-", "*" list markers
\item Markdown header frequency: Detects "\#", "\#\#," etc. patterns
\end{itemize}

\underline{Tonal Markers (3 features):}
\begin{itemize}
\item Hedging language: Frequency of "might", "could", "possibly", "perhaps," and "maybe" per 1000 tokens
\item Apologetic language: Frequency of "apologize" and "sorry" per 1000 tokens
\item Formal tone: Frequency of "furthermore", "moreover", "consequently", "nevertheless," and "thereby" per 1000 tokens
\end{itemize}

\subsection{Experiment 4.2 Details: Feature Extraction and Validation}
\subsubsection{Feature Extraction Implementation}
All features are extracted using deterministic rules implemented in Python with NLTK for sentence tokenization. Punctuation features use regular expression patterns with explicit Unicode character matching. Discourse markers use case-insensitive string matching with word boundary constraints to avoid substring matches. Structural elements use pattern matching with context validation, such as numbered lists require ascending sequences.

\subsubsection{Baseline Statistics Calculation}
For each corpus (Pile, Dolma), we compute:
\begin{itemize}
\item Mean $\mu_f$ and standard deviation $\sigma_f$ for each feature $f$
\item Inter-document variance to assess baseline diversity
\item CV to assess relative variability
\end{itemize}

These statistics are saved for consistent normalization across all experiments.

\subsubsection{Validation Metrics}
To ensure feature reliability, we assess:
\begin{enumerate}
\item \textbf{Test-retest reliability}: Extract features from same documents with 1-week interval. Pearson $r > 0.95$ confirms stable extraction.

\item \textbf{Baseline diversity}: Measure coefficient of variation across corpus samples. High baseline variance indicates natural human variation, validating that models amplify beyond normal ranges.
\end{enumerate}

\subsubsection{Baseline Statistics}

Feature extraction is highly reliable, with test–retest correlations of $r=0.997$ (Pile) and $r=0.999$ (Dolma). Both corpora exhibit substantial stylistic diversity (CV$_{mean}$ = 77.23 for the Pile, 53.26 for Dolma). Considerable baseline variation (CV $>$ 0.5 for 22/24 features for the Pile, 24/24 features for Dolma) indicates that model deviations are measured against a naturally variable human baseline rather than a narrow distribution. Consequently, observed amplification ratios and stylistic distortions in model outputs can be interpreted as genuine departures from typical human usage rather than artifacts of low-variance data. While absolute baseline frequencies differ between corpora, their overall magnitudes and relative rankings are stable, making the Pile a reliable reference for models whose training data is not directly accessible.

\clearpage
\begin{table}[ht]
\caption{Baseline feature frequencies ($\mu$) and variability ($\sigma$) across human-written corpora. Structural features (e.g., headers, lists) are rare in aggregate but exhibit high variance, reflecting strong context dependence, while local stylistic features (e.g., punctuation, tone) appear more consistently.}
\label{tab:baseline-stats}
\centering
\begin{tabular}{lcccc}
\toprule
\textbf{Feature} & \multicolumn{2}{c}{\textbf{The Pile}} & \multicolumn{2}{c}{\textbf{Dolma}} \\
\cmidrule(r){2-3} \cmidrule(l){4-5}
 & $\mu$ (\%) & $\sigma$ (\%) & $\mu$ (\%) & $\sigma$ (\%) \\
\midrule
\multicolumn{5}{l}{\textit{Punctuation}} \\
Em dashes & 0.038 & 2.42 & 0.074 & 1.18 \\
Semicolons & 0.876 & 19.78 & 0.140 & 1.16 \\
Colons & 0.528 & 7.78 & 0.233 & 2.04 \\
Ellipses & 0.072 & 3.66 & 0.143 & 1.42 \\
Parentheticals & 2.39 & 18.86 & 0.484 & 1.85 \\
\midrule
\multicolumn{5}{l}{\textit{Discourse Markers}} \\
"Delve into" & 0.00006 & 0.00 & 0.00042 & 0.026 \\
"It's worth noting" & 0.0 & 0.0 & 0.000021 & 0.0099 \\
"In conclusion" & 0.0014 & 0.036 & 0.00046 & 0.044 \\
"That being said" & 0.00024 & 0.017 & 0.00065 & 0.032 \\
"Arguably" & 0.00060 & 0.027 & 0.0013 & 0.10 \\
"Essentially" & 0.0031 & 0.097 & 0.0041 & 0.092 \\
"Fundamentally" & 0.00084 & 0.056 & 0.00086 & 0.034 \\
"Navigate" & 0.0036 & 0.84 & 0.0041 & 0.13 \\
"Landscape" & 0.0021 & 0.19 & 0.0042 & 0.12 \\
"Robust" & 0.0035 & 0.072 & 0.0024 & 0.073 \\
\midrule
\multicolumn{5}{l}{\textit{Sentence-Initial Markers}} \\
"However" (start) & 0.0054 & 0.17 & 0.0064 & 0.21 \\
"Certainly" & 0.00024 & 0.034 & 0.00015 & 0.071 \\
"Absolutely" & 0.00012 & 0.041 & 0.00033 & 0.088 \\
\midrule
\multicolumn{5}{l}{\textit{Structural Elements}} \\
Numbered lists & 0.058 & 4.84 & 0.027 & 1.04 \\
Bullet points & 0.112 & 3.27 & 0.0089 & 0.52 \\
Markdown headers & 0.034 & 2.13 & 0.000042 & 0.066 \\
\midrule
\multicolumn{5}{l}{\textit{Tonal Markers}} \\
Hedging language & 0.127 & 0.79 & 0.165 & 0.80 \\
Apologetic language & 0.0049 & 0.23 & 0.0040 & 0.33 \\
Formal tone & 0.036 & 0.44 & 0.016 & 0.18 \\
\bottomrule
\end{tabular}
\end{table}

\subsection{Prompt Templates}
For each model, we generate 1,000 text samples using the following protocol:

\underline{Prompts (15 diverse topics):}
\begin{enumerate}
\item "Write a detailed analysis of the benefits and drawbacks of remote work in modern society."
\item "Explain the complex relationship between technology and privacy in the digital age."
\item "Discuss the potential impacts of artificial intelligence on the job market over the next decade."
\item "Analyze the key factors contributing to climate change and potential solutions."
\item "Compare and contrast different approaches to education reform in the 21st century."
\item "Examine the role of social media in shaping public opinion and political discourse."
\item "Discuss the ethical implications of genetic engineering and CRISPR technology."
\item "Analyze the economic and social effects of globalization on developing nations."
\item "Explore the relationship between mental health and modern lifestyle factors."
\item "Discuss the challenges and opportunities of renewable energy adoption."
\item "Examine the impact of streaming services on traditional media industries."
\item "Analyze the factors that contribute to successful entrepreneurship in tech startups."
\item "Discuss the implications of automation and robotics on manufacturing industries."
\item "Explore the concept of work-life balance in contemporary professional culture."
\item "Analyze the role of regulation in cryptocurrency and blockchain technology."
\end{enumerate}

 Prompts are cycled to generate 1,000 samples, ensuring topic diversity while maintaining sufficient per-prompt sample size for robust statistics.

\subsection{Models}
We evaluate 13 models:

\underline{Open-Source Base Models (10 models):}
\begin{itemize}
\item EleutherAI/pythia-410m (410M parameters) \citep{biderman2023pythia}
\item allenai/OLMo-2-0425-1B (1B parameters) \citep{groeneveld2024olmo}
\item allenai/OLMo-2-0425-1B-Instruct (1B parameters, instruction-tuned) \citep{groeneveld2024olmo}
\item mistralai/Mistral-7B-v0.3 (7B parameters) \citep{jiang2023mistral}
\item mistralai/Mistral-7B-Instruct-v0.2 (7B parameters, instruction-tuned) \citep{jiang2023mistral}
\item meta-llama/Llama-3.2-3B (3B parameters) \citep{dubey2024llama}
\item meta-llama/Llama-3.2-3B-Instruct (3B parameters, instruction-tuned) \citep{dubey2024llama}
\item meta-llama/Llama-3.1-8B (8B parameters) \citep{dubey2024llama}
\item meta-llama/Llama-3.1-8B-Instruct (8B parameters, instruction-tuned) \citep{dubey2024llama}
\item google/gemma-2-2b-it (2B parameters, instruction-tuned) \citep{team2024gemma}
\end{itemize}
\underline{Commercial APIs (3 models):}
\begin{itemize}
\item gpt-4o-mini (OpenAI, frontier-scale) \citep{hurst2024gpt}
\item claude-3-5-haiku-20241022 (Anthropic, frontier-scale) \citep{anthropic2024claude}
\item gemini-2.5-flash (Google, frontier-scale) \citep{team2024gemini}
\end{itemize}

 This selection provides coverage across: (1) model sizes from 410M to 100B+ parameters, (2) training paradigms including base pretraining, instruction tuning, and RLHF, and (3) architectural families (Pythia, OLMo, Mistral, Llama, Gemma, GPT, Claude, Gemini).

\subsection{Experiment 4.3 Details: Model Infrastructure and Computation}
\subsubsection{Generation Parameters}
\begin{itemize}
\item Temperature: 0.7, balancing fluency and diversity
\end{itemize}

 Parameters are fixed across all models to isolate architectural and training effects from sampling variability.

\subsubsection{Computational Infrastructure}
Open-source models are evaluated on Lambda Labs 8$\times$A100 (40GB) instance. Generation uses HuggingFace Transformers 4.38.0 with torch 2.2.0 and CUDA 12.1. Commercial APIs are queried sequentially with rate-limit handling. All generations are deterministic, using seed=42, where supported.

\subsubsection{Feature Extraction}
For each model's 1,000 outputs, we extract all 24 features using the validated extraction pipeline from Section A.2. Features are aggregated as:
\begin{itemize}
\item Mean frequency across all outputs
\item Maximum frequency to identify outliers
\item Statistically significant features to capture consistency
\end{itemize}

\subsubsection{Divergence Calculation}
For each model-feature pair, we compute amplification ratio and aggregate divergence per model.

\subsubsection{Statistical Testing}
We validate measurement reliability through two approaches:

\begin{itemize}
\item \textbf{Sample size}: Each model generates 1,000 outputs, providing stable frequency estimates, with standard error $<$0.01\% for features with $>$1\% baseline frequency.

\item \textbf{Bonferroni-corrected significance}: To account for multiple comparisons across 24 features, we apply family-wise error control with $\alpha = 0.05 / 24 = 0.0022$. Feature subset analysis demonstrates that structural features achieve significant correlation with overall divergence ($\rho=0.560$, $p=0.0463$), validating that observed amplifications exceed measurement noise.

\item \textbf{Effect magnitude}: Amplification ratios directly quantify effect sizes (e.g., headers amplified by 16,753\% signifies an AR of 16,853\%), providing more interpretable measures than standardized effect sizes while maintaining comparability across features through percentage-based normalization.
\end{itemize}

\subsection{Experiment 4.4 Details: Entropy Regularization Training and Evaluation}
\subsubsection{Loss Function}

$$\mathcal{L} = \mathcal{L}_{CE} - \lambda \cdot H(P_\theta)$$

where $\mathcal{L}_{CE}$ is standard cross-entropy loss and $H(P_\theta) = -\sum_t P_\theta(x_t | x_{<t}) \log P_\theta(x_t | x_{<t})$ is the entropy of the output distribution.

\subsubsection{Training Configuration}
\begin{itemize}
\item Base architecture: Pythia-410M (24 layers, 16 attention heads, 1024 hidden dimension)
\item Training data: 40GB subset of The Pile (\(\sim 8\text{B tokens}\)), stratified across all constituent datasets
\item Batch size: 256 sequences $\times$ 2048 tokens = 524,288 tokens/batch
\item Learning rate: 6e-4 with cosine decay to 6e-5
\item Warmup: 1,000 steps (2\% of training)
\item Total steps: 50,000 (\(\sim 26\text{B tokens}\) seen, 3.25 epochs)
\item Optimizer: AdamW ($\beta_1=0.9$, $\beta_2=0.95$, weight decay=0.1)
\item Mixed precision: bf16 for memory efficiency
\end{itemize}

Training conducted on Lambda Labs 8$\times$A100 instance with DeepSpeed ZeRO Stage 2.

\subsubsection{Evaluation Protocol}
For each trained model, we generate 1,000 outputs using identical prompts as Section A.3 and identical parameters as Section A.5. We measure:

\begin{itemize}
\item \textbf{Stylistic divergence}: Aggregate $D$ across 23 features, feature-level amplification ratios
\item \textbf{Perplexity}: Held-out validation set (10M tokens from The Pile)
\item \textbf{Mode collapse}: Self-BLEU-4 (repetition), distinct-2/3/4 (diversity)
\end{itemize}

\subsubsection{Feature-Level Effects}

Having identified $\lambda=5.0$ as optimal, we evaluate its effectiveness through direct comparison with the unregularized baseline.

\begin{table}[H]
\centering
\caption{Feature-level effects comparing $\lambda$=0.0 vs $\lambda$=5.0. AR is amplification ratio. $\Delta$ (\%) = percentage change. $\Delta$ (pp) is additive change in percentage points. Key findings: (1) Non-monotonic effects: Regularization reduces some amplifications ("landscape" -82\%/-15.54pp; "essentially" -41\%/-0.245pp) but increases others (bullet points +0.639pp, "however" +0.847pp, hedging +139\%/+0.222pp). (2) Net improvement via large reductions: "Landscape" alone drops by 15.54pp, while all increases total only ~4pp, explaining overall divergence improvement (0.78 vs 0.63). (3) Restoration of suppressed features: Em dashes, headers, apologetic language, completely absent in $\lambda$=0.0, reappear (+0.125pp, +0.059pp, +0.028pp), demonstrating regularization restores distributional breadth.}
\small
\begin{tabular}{lccccc}
\toprule
\textbf{Feature} & \textbf{$\lambda=0.0$} & \textbf{$\lambda=5.0$} & \textbf{$\Delta$ (\%)} & \textbf{$\Delta$ (pp)} & \textbf{$p$-value} \\
 & \textbf{AR} & \textbf{AR} & & & \\
\midrule
\multicolumn{6}{l}{\textit{Amplified Features (Reduced by Regularization)}} \\
Landscape & 18.97 & 3.43 & -81.9 & -15.54 & $< 0.001$ \\
"Essentially" & 0.605 & 0.360 & -40.5 & -0.245 & $< 0.01$ \\
Semicolons & 0.102 & 0.083 & -18.7 & -0.019 & 0.455 \\
\midrule
\multicolumn{6}{l}{\textit{Amplified Features (Increased by Regularization)}} \\
Bullet points & 0.003 & 0.642 & — & +0.639 & $< 0.001$ \\
"However" (start) & 0.020 & 0.867 & — & +0.847 & $< 0.001$ \\
Hedging language & 0.160 & 0.382 & +138.8 & +0.222 & $< 0.001$ \\
Numbered lists & 1.72 & 3.21 & +86.6 & +1.49 & $< 0.001$ \\
"In conclusion" & 0.416 & 0.656 & +57.8 & +0.240 & 0.032 \\
Formal tone & 0.312 & 0.445 & +42.8 & +0.133 & 0.067 \\
"Robust" & 3.48 & 4.64 & +33.5 & +1.16 & 0.089 \\
Colon (mid-sentence) & 0.299 & 0.352 & +17.7 & +0.053 & 0.234 \\
\midrule
\multicolumn{6}{l}{\textit{Suppressed Features (Restored by Regularization)}} \\
Em dashes & 0.0 & 0.125 & — & +0.125 & $< 0.001$ \\
Header & 0.0 & 0.059 & — & +0.059 & $< 0.05$ \\
Apologetic & 0.0 & 0.028 & — & +0.028 & $< 0.01$ \\
\midrule
\multicolumn{6}{l}{\textit{Never Used (Both Models)}} \\
\multicolumn{6}{l}{Ellipsis, Parenthetical, "Delve into", "It's worth noting", "That being said",} \\
\multicolumn{6}{l}{"Arguably", "Fundamentally", "Navigate", "Certainly", "Absolutely"} \\
\bottomrule
\end{tabular}
\end{table}

\subsubsection{Cross-Model Comparison}
\paragraph{Tier 1: Similar-Scale Models (1-3B)}\hspace{0pt}%

\begin{table}[H]
\centering
\caption{Comparison with similar-scale models (1-3B parameters, 2-7$\times$ larger). Despite being significantly smaller, pythia-410m-$\lambda$-5.0 achieves 92.6-99.8\% better proximity to optimal stylistic naturalness (distance from 1.0) across all comparisons, demonstrating that training-time intervention with entropy regularization outweighs both scale and instruction tuning for stylistic quality. The OLMo-Instruct result is particularly striking—instruction tuning at 1B parameters (distance: 103.6) produces catastrophically worse stylistic patterns than entropy-regularized training at 410M (distance: 0.22), a 99.8\% improvement.}
\small
\begin{tabular}{lccc}
\toprule
\textbf{Model} & \textbf{Size} & \textbf{Distance from 1.0} & \textbf{vs. $\lambda$=5.0} \\
\midrule
pythia-410m-$\lambda$-5.0 & 410M & 0.22 & — \\
\midrule
OLMo-1B (base) & 1B & 7.08 & 96.9\% better \\
OLMo-1B-Instruct & 1B & 103.60 & 99.8\% better \\
Gemma-2-2b-it & 2B & 9.13 & 97.6\% better \\
Llama-3.2-3B (base) & 3B & 2.97 & 92.6\% better \\
Llama-3.2-3B-Instruct & 3B & 9.87 & 97.8\% better \\
\bottomrule
\end{tabular}
\end{table}

\paragraph{Tier 2: Larger Models (7-8B)}\hspace{0pt}%

\begin{table}[H]
\centering
\caption{Comparison with 7-8B parameter models (17-20$\times$ larger). Even at massive scale disadvantage, $\lambda$=5.0 achieves 88.7-98.1\% better proximity to optimal stylistic naturalness. These models represent the typical scale for open-source deployments, yet entropy regularization at 410M outperforms them comprehensively, suggesting that training procedure matters dramatically more than scale for stylistic quality.}
\begin{tabular}{lccc}
\toprule
\textbf{Model} & \textbf{Size} & \textbf{Distance from 1.0} & \textbf{vs. $\lambda$=5.0} \\
\midrule
pythia-410m-$\lambda$-5.0 & 410M & 0.22 & — \\
\midrule
Mistral-7B (base) & 7B & 1.94 & 88.7\% better \\
Mistral-7B-Instruct & 7B & 5.98 & 96.3\% better \\
Llama-3.1-8B (base) & 8B & 11.39 & 98.1\% better \\
Llama-3.1-8B-Instruct & 8B & 9.64 & 97.8\% better \\
\bottomrule
\end{tabular}
\end{table}

\subsection{Ablation Studies}

To validate design choices in our measurement framework and experimental design, we conduct systematic ablations testing alternative approaches.

\subsubsection{Feature Subset Analysis}

\paragraph{Objective}\hspace{0pt}%
To test whether all 24 features are necessary or if smaller subset suffices.

\paragraph{Experimental Design}\hspace{0pt}%
We compute divergence using feature subsets:
\begin{enumerate}
\item \textbf{Full taxonomy (proposed)}: All 24 features
\item \textbf{Top-10 discriminative}: Highest mean $|AR - 1.0|$ across 13 production models
\item \textbf{Category subsets}: Punctuation (5), discourse markers (8), structural (3), tonal (8)
\end{enumerate}

\paragraph{Results}\hspace{0pt}%

\begin{table}[ht]
\centering
\caption{Feature subset analysis. Spearman $\rho$ measures rank correlation with full divergence scores across 13 production models. Variance captured computed via $R^2$ from Pearson correlation. MAE is mean absolute error in divergence prediction. Top-10 features (42\% of taxonomy) achieve perfect rank correlation ($\rho=1.000$) and capture 100\% of variance, demonstrating that a small subset of highly discriminative features drives divergence patterns. These features span structural (headers, lists), discourse ("in conclusion", "however"), and tonal markers ("landscape", "navigate", "robust"), confirming that AI voice manifests across multiple linguistic dimensions rather than being dominated by any single category. Full taxonomy retained for comprehensiveness, model-specific patterns, and interpretability.}
\begin{tabular}{lcccc}
\toprule
\textbf{Subset} & \textbf{N Features} & \textbf{Spearman $\rho$} & \textbf{Variance (\%)} & \textbf{MAE} \\
\midrule
Full taxonomy & 24 & 1.000 & 100.0 & 0.000 \\
Top-10 & 10 & 1.000 & 100.0 & 19.824 \\
Structural only & 3 & 0.560 & 98.4 & 58.944 \\
Discourse only & 8 & 0.511 & 32.5 & 11.951 \\
Tonal only & 8 & 0.522 & 0.1 & 10.893 \\
Punctuation only & 5 & 0.401 & 11.8 & 14.581 \\
\bottomrule
\end{tabular}
\end{table}

\paragraph{Top-10 Most Discriminative Features (Mean Divergence):}\hspace{0pt}%

\begin{enumerate}
\item Headers: 16,853\%
\item "In conclusion": 5,048\%
\item "Delve into": 3,660\%
\item Bullet points: 3,063\%
\item "Landscape": 2,891\%
\item Numbered lists: 1,949\%
\item "Navigate": 958\%
\item "Robust": 684\%
\item "Fundamentally": 570\%
\item "However" (sentence-initial): 332\%
\end{enumerate}

These 10 features span structural (headers, lists), discourse ("in conclusion", "however"), and tonal markers ("landscape", "navigate", "robust"), confirming that AI voice manifests across multiple linguistic dimensions.

\subsubsection{Normalization Strategy}

\paragraph{Objective}\hspace{0pt}%
To validate percentage-based amplification ratios over alternatives.

\paragraph{Results}\hspace{0pt}%

\begin{table}[ht]
\centering
\caption{Normalization comparison. Percentage-based amplification ratios provide intuitive interpretation (e.g., "headers amplified to 16,853\%"), enable cross-feature aggregation via mean divergence, and appropriately weight features by baseline frequency, as rare features require larger absolute changes to achieve same percentage deviation.}
\begin{tabular}{lccc}
\toprule
\textbf{Normalization} & \textbf{Interpretability} & \textbf{Comparability} & \textbf{Sensitivity} \\
\midrule
Percentage (proposed) & High & Yes & Low \\
Z-score & Medium & Yes & High \\
Absolute & Low & No & Medium \\
\bottomrule
\end{tabular}
\end{table}

\subsubsection{Computational Efficiency}

\paragraph{Feature Extraction}\hspace{0pt}%

\begin{table}[ht]
\centering
\caption{Feature extraction performance measured on model outputs, with mean token length ~300. Time per document remains constant at ~0.3ms across corpus sizes, confirming linear $O(n)$ scaling. Full baseline construction of 10,000 documents completes in under 5 seconds on single CPU core, making framework highly efficient for large-scale analysis.}
\begin{tabular}{ccc}
\toprule
\textbf{Documents} & \textbf{Total Time} & \textbf{Time per Document} \\
\midrule
10 & 0.003 sec & 0.28 ms \\
50 & 0.015 sec & 0.30 ms \\
100 & 0.033 sec & 0.33 ms \\
500 & 0.158 sec & 0.32 ms \\
\bottomrule
\end{tabular}
\end{table}



\subsubsection{Summary}

Ablations validate framework design choices:

\begin{itemize}
\item \textbf{Feature coverage}: Top-10 features achieve perfect correlation ($\rho=1.000$) and 100\% variance capture, but full 24-feature taxonomy retained for comprehensive coverage of model-specific patterns and interpretability across diverse stylistic dimensions.

\item \textbf{Normalization}: Percentage-based amplification ratios maximize interpretability while enabling cross-feature aggregation through mean divergence metric.

\item \textbf{Efficiency}: Feature extraction completes in 0.3ms per document with linear scaling, and entropy regularization adds only 4\% training time, establishing both measurement and mitigation as computationally practical for large-scale deployment.
\end{itemize}


\newpage
\input{checklist.tex}

\end{document}

%% file: checklist.tex
\section*{NeurIPS Paper Checklist}

The checklist is designed to encourage best practices for responsible machine learning research, addressing issues of reproducibility, transparency, research ethics, and societal impact. Do not remove the checklist: {\bf The papers not including the checklist will be desk rejected.} The checklist should follow the references and follow the (optional) supplemental material.  The checklist does NOT count towards the page
limit. 

Please read the checklist guidelines carefully for information on how to answer these questions. For each question in the checklist:
\begin{itemize}
    \item You should answer \answerYes{}, \answerNo{}, or \answerNA{}.
    \item \answerNA{} means either that the question is Not Applicable for that particular paper or the relevant information is Not Available.
    \item Please provide a short (1--2 sentence) justification right after your answer (even for \answerNA). 
\end{itemize}

{\bf The checklist answers are an integral part of your paper submission.} They are visible to the reviewers, area chairs, senior area chairs, and ethics reviewers. You will also be asked to include it (after eventual revisions) with the final version of your paper, and its final version will be published with the paper.

The reviewers of your paper will be asked to use the checklist as one of the factors in their evaluation. While \answerYes{} is generally preferable to \answerNo{}, it is perfectly acceptable to answer \answerNo{} provided a proper justification is given (e.g., error bars are not reported because it would be too computationally expensive'' or ``we were unable to find the license for the dataset we used''). In general, answering \answerNo{} or \answerNA{} is not grounds for rejection. While the questions are phrased in a binary way, we acknowledge that the true answer is often more nuanced, so please just use your best judgment and write a justification to elaborate. All supporting evidence can appear either in the main paper or the supplemental material, provided in appendix. If you answer \answerYes{} to a question, in the justification please point to the section(s) where related material for the question can be found.

IMPORTANT, please:
\begin{itemize}
    \item {\bf Delete this instruction block, but keep the section heading ``NeurIPS Paper Checklist"},
    \item  {\bf Keep the checklist subsection headings, questions/answers and guidelines below.}
    \item {\bf Do not modify the questions and only use the provided macros for your answers}.
\end{itemize}


\begin{enumerate}

\item {\bf Claims}
    \item[] Question: Do the main claims made in the abstract and introduction accurately reflect the paper's contributions and scope?
    \item[] Answer: \answerYes{} 
    \item[] Justification: The abstract and introduction state six contributions, catastrophic divergence documentation, alignment training independence, selective amplification patterns, mechanistic framework, control strength principle, and cross-architecture validation, all of which are supported by experimental results in Sections 4.3-4.4 and the appendix.

\item {\bf Limitations}
    \item[] Question: Does the paper discuss the limitations of the work performed by the authors?
    \item[] Answer: \answerYes{} 
    \item[] Justification: Section 6 discusses limitations including the use of deterministic regex matching (prioritizing precision over recall), restriction to expository English prompts, confinement of entropy regularization experiments to Pythia-410M, and the lack of causal verification of the mechanistic account.

\item {\bf Theory assumptions and proofs}
    \item[] Question: For each theoretical result, does the paper provide the full set of assumptions and a complete (and correct) proof?
    \item[] Answer: \answerNA{} 
    \item[] Justification: The paper presents a mechanistic framework (Section 3) with formal definitions and hypotheses rather than formal theorems requiring proofs.

    \item {\bf Experimental result reproducibility}
    \item[] Question: Does the paper fully disclose all the information needed to reproduce the main experimental results of the paper to the extent that it affects the main claims and/or conclusions of the paper (regardless of whether the code and data are provided or not)?
    \item[] Answer: \answerYes{} 
    \item[] Justification: The paper provides complete experimental details: feature taxonomy and detection methods (Section 4.1, Appendix A.1), baseline construction procedure (Section 4.1.2, Appendix A.2), all 15 prompt templates (Appendix A.3), full model list with versions (Appendix A.4), generation parameters (Appendix A.5), training configuration for entropy regularization including architecture, data, hyperparameters, and optimizer settings (Appendix A.6), and evaluation protocol.

\item {\bf Open access to data and code}
    \item[] Question: Does the paper provide open access to the data and code, with sufficient instructions to faithfully reproduce the main experimental results, as described in supplemental material?
    \item[] Answer: \answerYes{} 
    \item[] Justification: Code used for experimentation is included as supplementary material and will be released publicly upon acceptance.

\item {\bf Experimental setting/details}
    \item[] Question: Does the paper specify all the training and test details (e.g., data splits, hyperparameters, how they were chosen, type of optimizer) necessary to understand the results?
    \item[] Answer: \answerYes{} 
    \item[] Justification: Generation parameters are specified in Appendix A.5 (temperature 0.7, max length 1024, seed=42). Training configuration for entropy regularization is fully detailed in Appendix A.6, including architecture (Pythia-410M), training data (40GB subset of The Pile), batch size, learning rate schedule, optimizer (AdamW with specified betas and weight decay), total steps, and mixed precision settings.

\item {\bf Experiment statistical significance}
    \item[] Question: Does the paper report error bars suitably and correctly defined or other appropriate information about the statistical significance of the experiments?
    \item[] Answer: \answerYes{} 
    \item[] Justification: The paper reports $p$-values for the base-instruct comparison (Table 3, all $p >$ 0.25), provides baseline feature frequencies $\mu$ and variability $\sigma$, applies Bonferroni correction for multiple comparisons across 24 features ($\alpha = 0.05/24 = 0.0022$, Appendix A.5.5), reports Spearman correlation with $p$-values for scale-divergence analysis ($\rho = 0.21$, $p = 0.49$), and provides feature-level $p$-values in Table 8. Standard error bounds for baseline frequencies are also reported ($<$ 0.000032 for features with baseline $\geq$ 0.01\%) along with Spearman correlation $\rho$, variance, and MAE in feature subset analysis (Appendix A.7.1).

\item {\bf Experiments compute resources}
    \item[] Question: For each experiment, does the paper provide sufficient information on the computer resources (type of compute workers, memory, time of execution) needed to reproduce the experiments?
    \item[] Answer: \answerYes{} 
    \item[] Justification: Appendix A.5.2 specifies the computational infrastructure: Lambda Labs 8$\times$A100 (40GB) instance for open-source model evaluation and entropy regularization training, with software versions (HuggingFace Transformers 4.38.0, torch 2.2.0, CUDA 12.1), and commercial APIs were queried sequentially with rate-limit handling. Training uses DeepSpeed ZeRO Stage 2, and bf16 was used for memory efficiency (Appendix A.6.2). Execution times for feature extraction are documented in Appendix A.7.3.
    
\item {\bf Code of ethics}
    \item[] Question: Does the research conducted in the paper conform, in every respect, with the NeurIPS Code of Ethics \url{https://neurips.cc/public/EthicsGuidelines}?
    \item[] Answer: \answerYes{} 
    \item[] Justification: The research uses publicly available models and corpora, does not involve human subjects, and discusses potential dual-use concerns transparently.

\item {\bf Broader impacts}
    \item[] Question: Does the paper discuss both potential positive societal impacts and negative societal impacts of the work performed?
    \item[] Answer: \answerYes{} 
    \item[] Justification: Section 5 (Discussion) and Section 7 (Conclusion) discuss positive impacts (understanding and correcting distributional collapse, improving linguistic diversity in AI outputs) and negative impacts (entropy-regularized models could evade AI detection systems, implications for training data contamination and long-term linguistic evolution).
    
\item {\bf Safeguards}
    \item[] Question: Does the paper describe safeguards that have been put in place for responsible release of data or models that have a high risk for misuse (e.g., pre-trained language models, image generators, or scraped datasets)?
    \item[] Answer: \answerNA{} 
    \item[] Justification: The paper does not release pre-trained models or scraped datasets. The entropy-regularized Pythia-410M models are small-scale research artifacts that do not pose risks beyond those of the publicly available base Pythia models.

\item {\bf Licenses for existing assets}
    \item[] Question: Are the creators or original owners of assets (e.g., code, data, models), used in the paper, properly credited and are the license and terms of use explicitly mentioned and properly respected?
    \item[] Answer: \answerYes{} 
    \item[] Justification: All models are cited by their original publications or model cards. The Pile is used under its public research license. Dolma is used under its AI2 license. HuggingFace Transformers and DeepSpeed are open-source under Apache 2.0. Commercial APIs (OpenAI, Anthropic, Google) are used in accordance with their terms of service.

\item {\bf New assets}
    \item[] Question: Are new assets introduced in the paper well documented and is the documentation provided alongside the assets?
    \item[] Answer: \answerNA{} 
    \item[] Justification: The paper does not release new datasets or models as standalone assets. The experimentation code is provided as supplemental material for reproducibility, not as independently released assets.

\item {\bf Crowdsourcing and research with human subjects}
    \item[] Question: For crowdsourcing experiments and research with human subjects, does the paper include the full text of instructions given to participants and screenshots, if applicable, as well as details about compensation (if any)? 
    \item[] Answer: \answerNA{} 
    \item[] Justification: The paper does not involve crowdsourcing or research with human subjects.

\item {\bf Institutional review board (IRB) approvals or equivalent for research with human subjects}
    \item[] Question: Does the paper describe potential risks incurred by study participants, whether such risks were disclosed to the subjects, and whether Institutional Review Board (IRB) approvals (or an equivalent approval/review based on the requirements of your country or institution) were obtained?
    \item[] Answer: \answerNA{} 
    \item[] Justification: The paper does not involve crowdsourcing or research with human subjects.

\item {\bf Declaration of LLM usage}
    \item[] Question: Does the paper describe the usage of LLMs if it is an important, original, or non-standard component of the core methods in this research? Note that if the LLM is used only for writing, editing, or formatting purposes and does \emph{not} impact the core methodology, scientific rigor, or originality of the research, declaration is not required.
    \item[] Answer: \answerNA{} 
    \item[] Justification: LLMs were used as assistive tools for code generation, editing, and some statistics formalization during the research process but did not contribute to the core methodology, experimental design, theoretical framework, or scientific conclusions of this work.

\end{enumerate}